# A Comparison of Bounding Box and Landmark Detection Methods for Video-Based Heart Rate Estimation

**Laurence Liang**
laurence.liang@mail.mcgill.ca
McGill University, Inspire

**Abstract**

Remote Photoplethysmography (rPPG) uses the cyclic variation of skin tone on a person's forehead region to estimate that person's heart rate. This paper compares two methods: a bounding box-based method and a landmark-detection-based method to estimate heart rate, and discovered that the landmark-based approach has a smaller variance in terms of model results with a standard deviation that is more than 4 times smaller (4.171 compared to 18.720).

## 1. Introduction

Remote photoplethysmography (rPPG) is a method that estimates features such as a person's heart rate by measuring the intensity of certain colours from a specified skin region. For heart rate, one approach that has been studied in the 2010s and early 2020s involves tracking the variability of green pixels in a person's forehead region.

One advantage of video-based rPPG compared to contact-based pulse monitoring devices is that rPPG could be used with patients who have sensitive skin, and in emergency situations where it is not possible or highly inconvenient to attach physical sensors to a patient.

## 2. Methods

This paper uses remote rPPG to contactlessly estimate the heart rate of a person through a video recording. The objective is to capture the forehead region above the participant's eyebrows, and monitor the variation of green light in that region, as the tiny, cyclic variations in green should correlate to the participant's heart rate. [1]

### 2.1 Setting

The 720p camera captures a 10 second video (± 1 second) of a seated person's face in a static environment with a white background. In this paper, there is only a single human participant.

We then record 10 videos of the seated person consecutively while in the same position.

### 2.2 Bounding Box Method

We use the BlazeFace model through MediaPipe in order to detect and track the coordinates of the participant's face in the recorded videos. [2] Next, we defined 4 coordinates as the corners of the bounding box that correspond to the forehead region of interest. We then extract the average values of green from that rectangular bounding box region.

We then apply a fast Fourier transform, and only consider frequencies between $f = 1.0\ Hz$ and $f = 4.0\ Hz$, as those regions would correspond to the expected human heart rate that should fall between 60 and 240 beats per minute.

### 2.3 Landmark Method

We use the Face Landmarker model through MediaPipe to track the forehead region of interest. We then apply a mask to track the pixels of interest from one frame to the next, and then extract the average values of green from that region.

We then apply a fast Fourier transform with the same methodology as (2.2) for the Bounding Box Method.

## 3. Results

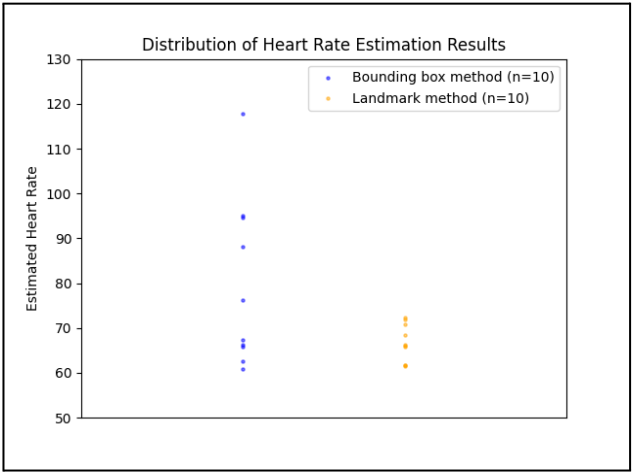

*Figure 1: Distribution of Heart Rate Estimation Methods*

Table 1: Mean and Standard Deviation of the Results from Bounding Box and Landmark Methods

| Statistic | Bounding Box Method (n=10) | Landmark Method (n=10) |
|---|---|---|
| Mean | 79.472 | 66.660 |
| Stdev | 18.720 | 4.171 |

We observe that the results obtained from the Bounding Box Method have a much higher standard deviation (18.720) compared to the Landmark Method (4.171). The Landmark Method does not have any extreme outliers, as all its results fall between the range of 61 to 73 beats per minute. While a subset of the Bounding Box Method's results also fall within this range, multiple outlier results give values higher than 85 beats per minute, which skews the Bounding Box Method mean and the standard deviation.

## 4. Conclusion & Discussion

This paper suggests that in a static environment, the Landmark Method has a noticeably smaller standard deviation than the Bounding Box Method for estimating heart rate via rPPG. However, there are several limitations in this paper that need to be addressed in order to provide a more comprehensive evaluation of the Bounding Box Method and the Landmark Method.

The data collection process did not include a truth value measurement of the participant's heart rate. While this paper was able to compare the variability of both rPPG methods, this paper was not able to evaluate the average error for each measurement. The heart rate recorded by counting pulses by hand in the right carotid artery under the participant's neck over 10 seconds gave an estimate of 72 beats per minute 5 to 7 minutes after the video recordings. However, because the participant moved in between the video recordings and this measurement, the measured heart rate by counting pulses from the carotid artery may not be the most representative measurement of the truth value during the video recordings.

Additionally, all the 10 recordings were done on a single human participant, and having a large, representative group to record additional data would be required to develop a more comprehensive evaluation of the variability of both methods to contactlessly estimate heart rate.

All the recordings took place in a static environment with the same background and lighting. Future studies should evaluate videos recorded in different lighting environments as well as consider participants who are moving in order to provide video data that is more reflective of the dynamic environment in a real-world setting.

## 5. Acknowledgements

We would like to thank Kristina Kupferschmidt for her support and recommendations on how to establish a ground truth for heart rate estimation.